%% file: main.tex
\documentclass{article}
\usepackage{spconf,amsmath,amsfonts,graphicx}
\usepackage{hyperref}
\usepackage{multirow}
\usepackage{enumitem}
\setlist{nosep, leftmargin=14pt}

\usepackage{mwe} 

\title{Multivariate Wasserstein Functional Connectivity\\for Autism Screening}
\name{Oleg Kachan$^{1,2}$ \quad Alexander Bernstein$^{3}$}
\address{$^1$ Laboratory for Algebraic Topology and Applications, HSE Univeristy\\
$^2$ Neuro Center, Skolkovo Institute of Science and Technology\\
$^3$ AI Center, Skolkovo Institute of Science and Technology}
\begin{document}
%
\maketitle

\input{sections/abstract}

\input{sections/introduction}
\input{sections/methods}
\input{sections/results}
\input{sections/discussion}

\input{sections/ethics}


\input{main.bbl}

\end{document}

%% file: sections/abstract.tex
\begin{abstract}
Most approaches to the estimation of brain functional connectivity from the functional magnetic resonance imaging (fMRI) data rely on computing some measure of statistical dependence, or more generally, a distance between univariate representative time series of regions of interest (ROIs) consisting of multiple voxels. However, summarizing a ROI's multiple time series with its mean or the first principal component (1PC) may result to the loss of information as, for example, 1PC explains only a small fraction of variance of the multivariate signal of the neuronal activity.

We propose to compare ROIs directly, without the use of representative time series, defining a new measure of multivariate connectivity between ROIs, not necessarily consisting of the same number of voxels, based on the Wasserstein distance. We assess the proposed Wasserstein functional connectivity measure on the autism screening task, demonstrating its superiority over commonly used univariate and multivariate functional connectivity measures.
\end{abstract}

%% file: sections/introduction.tex
\section{Introduction}






Brain and its activity is a complex network consisting of neuronal populations, coactivity of which have to be consistent across large populations with individual differences and dependence on variables of interest such as age, socioeconomic status, and presence or absence of various neurodevelopmental and neurodegenerative deceases and conditions \cite{Geerligs2016}. This activity known as the functional connectivity (FC), opposed to the structural connectivity which is the actual physical network of the neuronal populations and their interconnections, is estimated from the resting state functional magnetic resonance imaging (rs-fMRI).

Deceases and conditions are shown to be related to the increased, decreased, abnormal or generally altered functional connectivity, with the autism spectrum disorder (ASD) being prominent example. ASD is a condition related to the brain development, which affects the individual's social behavior. ASD typically develops in infancy and early childhood and screened by the behavioral tests which may be subjective. Therefore, it is important to bring the quantitative, interpretable methods to screen autism, with fMRI being the promising modality to address that purpose.

The common approach to estimate FC consist of dividing the raw fMRI image time series into ROI groups defined by parcellation atlas. Next the representative time series of each ROI are obtained by averaging or principal component analysis (PCA), and some measure of dependence between each pair of representatives, most commonly the Pearson correlation coefficient is computed. We refer to this pipeline as the univariate approach, based on univariate time series representatives, assuming the homogeneity of the time series in each ROI.

Although as commonly raw time series are grouped into ROI based not on their similarity, but according to atlases based on neuroanatomy or brain's cytoarchitecture, one could not expect time series homogeneity within a ROI or single mode of their distribution. Recently introduced multivariate approach \cite{Geerligs2016,Walther2016,Yoo2019,Basti2020} addresses this issue, with the most popular test for multivariate independence characterizing the relation strength is the distance correlation \cite{Szekely2007}. 

In our work we propose and assess a new multivariate measure of the functional connectivity based on the Wasserstein distance \cite{Kantorovich1958,Vaserstein1969} with the application to ASD screening.


%% file: sections/methods.tex
\section{Materials and methods}


\subsection{Functional connectivity}

The functional connectivity is estimated from fMRI data using some measure of dependence between time series of each ROI, most often Pearson correlation coefficient between the represetative time series of each ROI obtained by averaging or taking the first principal component.

More formally, given $\{X_i\}_{i=1}^N$ random variables with equal number of observations $t$, but of different dimensions $\{n_i\}_{i=1}^N$ where $N$ is the number of ROIs, one seeks to estimate some measure of statistial dependence $\rho_{ij}$ for each pair $(X_i, X_j)$. If such estimating is done using $1$-dimensional summaries of each $X_i$ we call that \emph{univariate} approach, otherwise if $X_i$ and $X_j$ are compared directly we call that \emph{multivariate} approach.



\input{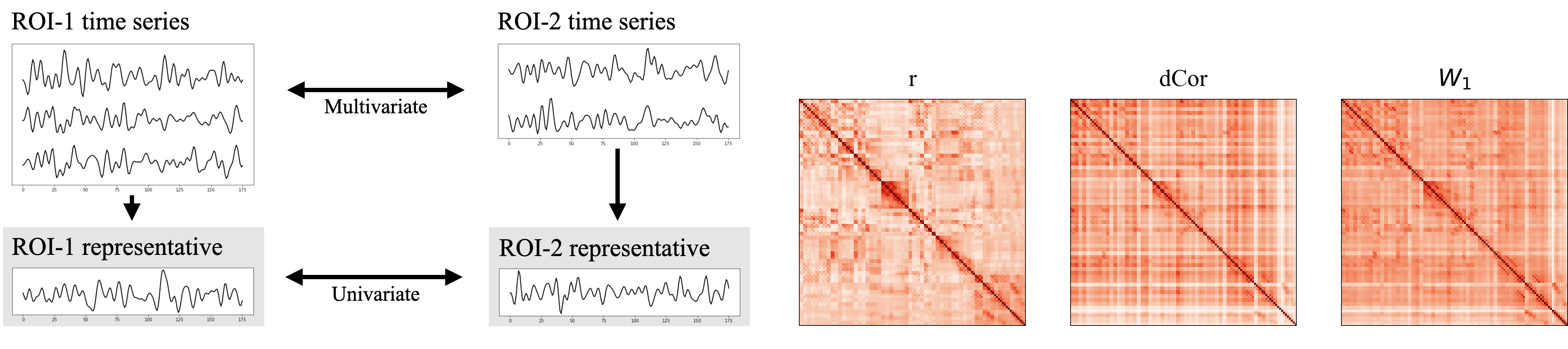}

\subsubsection{Pearson correlation}

Most often used similarity metric for comparing fMRI time series is \emph{Pearson correlation}, which is defined for population as

\vspace{-1em}

\begin{equation}
\rho(X, Y) = \frac{\mathrm{Cov}(X, Y)}{\sqrt{\mathrm{Var}(X) \mathrm{Var}(Y)}}.
\end{equation}
    


Empirical Pearson correlation is given by

\vspace{-0.5em}

\begin{align}
    r(\mathbf{x}, \mathbf{y})& = \frac{\sum_{i=1}^n (x_i - \bar{\mathbf{x}}) (y_i - \bar{\mathbf{y}})}{\sqrt{\sum_{i=1}^n x_i^2}\sqrt{\sum_{i=1}^n y_i^2}}\\
    &= \frac{\langle \mathbf{x}, \mathbf{y} \rangle}{\|\mathbf{x}\|_2\|\mathbf{y}\|_2},
\end{align}

with (3) valid for z-transformed vectors, standardized to the zero mean and unit variance.

\subsubsection{Distance correlation}

Pearson's correlation coefficient captures only linear association, with rank correlation coefficients traditionally used to address this issue. The relatively recently introduced distance correlation \cite{Szekely2007} in addition to being able to capture nonlinear dependencies is a multivariate measure of association between the groups of random variables of arbitrary, not of necessarily equal size. The idea of distance correlation is based on measuring the correlation of $t^2$ pairwise distances between realizations of random variables $X \in \mathbb{R}^n$ and $Y \in \mathbb{R}^m$, with population \emph{distance correlation} is defined analogously to the Pearson correlation

\begin{equation}
    dCor(X, Y)^2 = \frac{\mathrm{dCov}^2(X, Y)}{\sqrt{\mathrm{dVar}^2(X) \mathrm{dVar}^2(Y)}}.
\end{equation}


For an observed random sample, given by the ROI matrices $\mathbf{X}, \mathbf{Y} = \{(X_k, Y_k) \mid 1, \dots, n \}$ from the joint distribution of random vectors $X \in \mathbb{R}^p$ and $Y \in \mathbb{R}^q$ define

\begin{align}
    &a_{kl} = |X_k - X_l|_p,\nonumber\\
    \bar{a}_{k\cdot} = \frac{1}{n} \sum_{l=1}^n a_{kl},~~~~~~&\bar{a}_{\cdot k} = \frac{1}{n} \sum_{k=1}^n a_{kl},&\bar{a}_{\cdot\cdot} = \frac{1}{n^2} \sum_{k,l=1}^n a_{kl}\nonumber
\end{align}

$k,l = 1, \dots, n.$ to construct double centered pairwise distance matrix $\mathbf{A}$ (and simirarly for $\mathbf{B}$) with entries given by

\begin{equation}
    (\mathbf{A})_{kl} = a_{kl} - \bar{a}_{k\cdot} - \bar{a}_{\cdot k} + \bar{a}_{\cdot\cdot}.
\end{equation}

The empirical distance covariance $dCov_n(\mathbf{X}, \mathbf{Y})$ is defined by

\begin{equation}
    dCov_n^2(\mathbf{X}, \mathbf{Y}) = \frac{1}{n^2} \sum_{k,l=1}^n A_{kl}B_{kl}.
\end{equation}

The empirical distance variance $dVar_n(\mathbf{X})$ is defined by

\begin{equation}
    dVar_n^2(\mathbf{X}) = \frac{1}{n} \sum_{k,l=1}^n A_{kl}^2.
\end{equation}

Finally, the empirical \emph{distance correlation} $dCor_n(\mathbf{X}, \mathbf{Y})$ is defined

\vspace{-1em}

\begin{equation}
    dCor_n(\mathbf{X}, \mathbf{Y}) = \frac{dCov_n^2(\mathbf{X})}{\sqrt{dVar_n^2(\mathbf{X})dVar_n^2(\mathbf{Y})}},
\end{equation}

for $dVar_n^2(\mathbf{X})dVar_n^2(\mathbf{Y}) > 0$ and zero otherwise.

\subsubsection{Wasserstein distance}

Optimal transport considers comparing the measures over the domain $X$. Given a ground metric $d: X \times X \rightarrow \mathbb{R}$ optimal transport equips the space of measures $\mathcal{P}(X)$ with a metric referred to as the \emph{Wasserstein distance} \cite{Kantorovich1958,Vaserstein1969}, which for any $\mu, \nu \in \mathcal{P}(X)$ and $p \geq 1$ is defined as \cite{Peyre2019}

\begin{equation}
    W_p^p(\mu, \nu) = \inf_{\pi \in \Pi(\mu, \nu)} \int_{X \times X} d^p(x, y) d \pi(x, y)
\end{equation}

where $W_p^p$ denotes the $p$-th power of $W_p$ and $\Pi(\mu, \nu)$ is the set of probability measures on the product space $X \times X$ whose marginals coincide with $\mu$ and $\nu$; namely

\vspace{-1.25em}

\begin{equation*}
    \small
    \Pi(\mu, \nu) = \{ \pi \in \mathcal{P}(X \times X) \mid P_1\#\pi = \mu, P_2\#\pi = \nu \}.
\end{equation*}



Given ROI matrices $\mathbf{X} \in \mathbb{R}^{n \times t}$ and $\mathbf{Y} \in \mathbb{R}^{m \times t}$, consider the discrete measures $\mu_{\mathbf{X}}, \nu_{\mathbf{Y}} \in \mathcal{P}(X)$ that can be written as linear combinations $\mu_{\mathbf{X}} = \sum_{i=1}^n a_i \delta_{\mathbf{x}_i}$ and $\nu_{\mathbf{Y}} = \sum_{j=1}^m b_j \delta_{\mathbf{y}_j}$ of Dirac's deltas centred at a finite number $n$ and $m$ of points $(\mathbf{x}_i)_{i=1}^n \in \mathbb{R}^t$ and $(\mathbf{y}_j)_{j=1}^m \in \mathbb{R}^t$. In order for $\mu$ and $\nu$ to be probabilities, set the vector weights $\mathbf{a} = (1/n,\dots,1/n)^T \in \mathbb{R}^n$ and $\mathbf{b} = (1/m,\dots,1/m)^T \in \mathbb{R}^m$.

We define the \emph{Wasserstein functional connectivity (WFC)} as the Wasserstein distance between the two discrete measures $\mu_{\mathbf{X}}$ and $\nu_{\mathbf{Y}}$ with corresponding weight vectors $\mathbf{a}$ and $\mathbf{b}$ corresponds to

\begin{equation}
    W_p^p(\mu_{\mathbf{X}}, \nu_{\mathbf{Y}}) = \min_{\mathbf{T} \in \Pi(\mathbf{a}, \mathbf{b})} \langle \mathbf{T}, \mathbf{M} \rangle_F,
\end{equation}

where $\mathbf{M} \in \mathbb{R}^{n \times m}$ is the cost matrix with entries $(\mathbf{M})_{ij} = d^p(\mathbf{x}_i, \mathbf{y}_j)$ are set to the $\ell^p$-distance $d^p(\cdot, \cdot)$, $\langle \mathbf{T}, \mathbf{M} \rangle_F$ is the elementwise Frobenius inner product, and $\Pi(\mathbf{a}, \mathbf{b})$ denotes the transportation polytope

\vspace{-1.25em}

\begin{equation*}
    \Pi(\mathbf{a}, \mathbf{b}) = \{ \mathbf{T} \in \mathbb{R}^{n \times m}_+: \mathbf{T}\mathbf{1}_m = \mathbf{a}, \mathbf{T}^T\mathbf{1}_n = \mathbf{b} \}.
\end{equation*}



\subsection{Dataset}

Autism Brain Imaging Data Exchange (ABIDE) \cite{DiMartino2014} is a collaboration of $16$ international imaging sites that have aggregated and are openly sharing neuroimaging data. The dataset is composed of structural and resting state functional MRI (rs-fMRI) data along with an extensive array of phenotypic information. We used the preprocessed version\footnote{http://preprocessed-connectomes-project.org/abide/} with $879$ subjects total, with $474$ individuals suffering from ASD and $405$ typical controls. Functional preprocessing was performed using the Configurable Pipeline for the Analysis of Connectomes (CPAC) including the band-pass filtering and global signal regression.


\subsection{Experiment}


Functional connectivity matrices were computed using the Pearson correlation between the ROI representative time series obtained by averaging, and the distance correlation in the voxel space and the Wasserstein distance in the time space for full ROIs. Being symmetric with the same values on the diagonal, matrices were further vectorized by taking the upper half triangle for each matrix.

Following the best practices in fMRI data analysis described in Dadi et al. \cite{Dadi2019}, we used logistic regression with L$1$ and L$2$ regularization with the grid search for the regularization parameter on the $[0.2, 5]$ range. The data was split into the train and test subsets with the $80/20$ ratio, with the random downsampling of the largest class performed to mitigate the class imbalance. The classification accuracy mean and standard deviation in percents for the best pair of the regularization norm and regularization parameter value averaged over $20$ runs are reported in the Table \ref{table:results}.

Also we conducted the analysis of how considered connectivity measures are correlated between themselves. For that, first, mean connectivity matrices were computed for each measure, averaged for all subjects in the dataset. Second, the absolute differences between mean connectivity matrices of the ASD-affected and control patients were computed. Then, for each set of matrices the correlation was computed for each pair in the set, summarizing the values for each pair in the Table \ref{table:correlations}.

%% file: figures/figure1.tex
\begin{figure*}[h!]
\centering
\begin{center}
    \includegraphics[width=\textwidth]{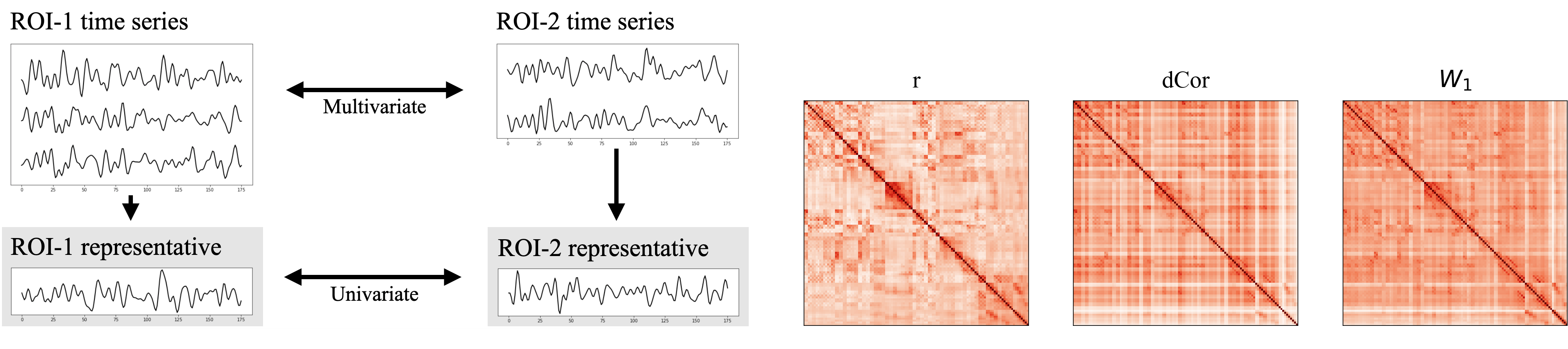}
\end{center}

\vspace{-2em}

\caption{\textbf{Left:} Computation of the functional connectivity. In the more common univariate approach a representative time series are computed for each ROI, for example by averaging or PCA and are compared. In the multivariate approach ROIs are compared directly, for example by distance correlation \cite{Szekely2007} or by the Wasserstein distance as in our proposal. \textbf{Right:} Mean connectivity matrices given by the Pearson and distance correlation and the Wasserstein distance. For the visualization matrices were transformed to have ones on the diagonal to facilitate the comparison.}
\label{figure:figure1}

\vspace{-1em}

\end{figure*}

%% file: sections/results.tex
\section{Results}

\vspace{-0.75em}

\input{tables/table1_all}

By the experimental evaluation on the ABIDE dataset -- the benchmark for autism diagnosis, we obtained the following results:
\begin{itemize}
    \item WFC outperforms the univarate and multivariate methods (Table \ref{table:results}) for the whole dataset composed of the data from all $16$ sites.
    \item WFC is competitive sitewise, quantified by the best average rank, including being ranked first on the $3$ largest sites -- NYU (171 subjects), UM (82), USM (61). For the Stanford site using the WFC metric leads to mean accuracy increase of $38.19\%$ compared to the Pearson correlation.
    \item WFC is more similar to dCor than to Pearson r, as shown with the visual inspection (Fig. \ref{figure:figure1}, right) and high correlation between them obtained for the averaged connectivity matrices as reflected in the Table \ref{table:correlations}, despite WFC is being defined in the time space, not the voxel space contrary to dCor.
\end{itemize}


`



\vspace{-0.75em}
\input{tables/table_correlations}

%% file: tables/table1_all.tex
\begin{table}[h!]
\centering
\begin{tabular}{|p{1.5cm}lll|} \hline
                               & \multicolumn{1}{c}{\small Univariate} & \multicolumn{2}{c|}{\small Multivariate}                 \\
                               & \multicolumn{1}{c}{Pearson r}       & \multicolumn{1}{c}{dCor} & \multicolumn{1}{c|}{Wasserstein} \\ \hline \hline
\multicolumn{1}{|l|}{All sites}&64.09$\pm$0.00 &62.95$\pm$0.00 &\textbf{65.54$\pm$0.00}\\\hline \hline 



\multicolumn{1}{|l|}{{Caltech}}&~56.02$\pm$8.84 &~54.37$\pm$7.33 &~\textbf{63.82$\pm$4.31}~\\
\multicolumn{1}{|l|}{{KKI}}&~55.40$\pm$11.87 &~\textbf{65.30$\pm$8.87} &~55.65$\pm$10.36~\\
\multicolumn{1}{|l|}{{Leuven}}&~59.05$\pm$5.58 &~\textbf{68.60$\pm$5.37} &~59.42$\pm$9.02~\\
\multicolumn{1}{|l|}{{MaxMun}}&~55.88$\pm$11.17 &~\textbf{80.14$\pm$4.23} &~63.18$\pm$7.10~\\
\multicolumn{1}{|l|}{{\textbf{NYU}}}&~62.96$\pm$3.35 &~58.51$\pm$3.02 &~\textbf{64.32$\pm$3.68}~\\
\multicolumn{1}{|l|}{{OHSU}}&~53.30$\pm$8.46 &~50.35$\pm$10.59 &~\textbf{53.75$\pm$11.63}~\\
\multicolumn{1}{|l|}{{Olin}}&~\textbf{68.60$\pm$7.74} &~60.35$\pm$13.45 &~63.60$\pm$9.52~\\
\multicolumn{1}{|l|}{{Pitt}}&~70.67$\pm$3.70 &~66.92$\pm$5.89 &~\textbf{73.04$\pm$5.41}~\\
\multicolumn{1}{|l|}{{SBL}}&~60.55$\pm$6.73 &~48.55$\pm$8.16 &~\textbf{68.00$\pm$12.08}~\\
\multicolumn{1}{|l|}{{SDSU}}&~\textbf{59.95$\pm$12.13} &~55.10$\pm$9.02 &~57.60$\pm$14.46~\\
\multicolumn{1}{|l|}{{Stanford}}&~58.88$\pm$6.12 &~93.57$\pm$2.95 &~\textbf{97.07$\pm$0.17}~\\
\multicolumn{1}{|l|}{{Trinity}}&~\textbf{64.29$\pm$8.14} &~51.85$\pm$6.01 &~56.57$\pm$6.42~\\
\multicolumn{1}{|l|}{{UCLA}}&~64.74$\pm$3.36 &~50.04$\pm$5.29 &~\textbf{65.34$\pm$3.05}~\\
\multicolumn{1}{|l|}{{\textbf{UM}}}&~68.65$\pm$5.20 &~70.41$\pm$4.57 &~\textbf{72.98$\pm$5.47}~\\
\multicolumn{1}{|l|}{{\textbf{USM}}}&~65.66$\pm$4.27 &~64.43$\pm$6.46 &~\textbf{69.81$\pm$5.21}~\\
\multicolumn{1}{|l|}{{Yale}}&~\textbf{69.43$\pm$9.02} &~56.35$\pm$6.26 &~64.76$\pm$8.04~\\ \hline

\multicolumn{1}{|l|}{{Best}}&4/16 &3/16 &\textbf{9/16} \\
\multicolumn{1}{|l|}{{Avg. rank}}&2.06 &2.50 &\textbf{1.44} \\ \hline
\end{tabular}
\caption{Comparison of the multivariate Wasserstein functional connectivity measure to the univariate and multivariate measures, represented by the Pearson and distance correlations respectively. Best classification accuracy, \% for the ASD/control classification task is reported for the $5$-fold cross-validation with $20$ repeats on the ABIDE1 dataset. The three largest sites are emphasized by the \textbf{bold} font.}
\label{table:results}
\end{table}

%% file: tables/table_correlations.tex
\begin{table}[h!]
\centering
\begin{tabular}{|p{2.2cm}|lll|} \hline
           &~~\small r/dDcor~& \small r/WFC   & \small dCor/WFC \\ \hline \hline
Mean       &~~0.265  ~& -0.584~~~& -0.893 \\
Difference &~~0.009  ~& 0.116~~~& 0.398 \\ \hline
\end{tabular}
\caption{Correlations between mean connectivity matrices computed using different connectivity measures and differences between ASD and control mean matrices.}
\label{table:correlations}
\end{table}

%% file: sections/discussion.tex
\vspace{-1em}
\section{Discussion}

Albeit shown to be beneficial for the downstream tasks, the main limitation of the Wasserstein distance connectivity measure is the increased computational complexity compared to both Pearson and distance correlations. This limitation can be addressed with several complementary approaches.

First is the use of dimensionality reduction -- we observed that applying PCA on the time domain, taking $25$\% principal components with the subsequent computation of the Wasserstein distance does affect the value of WFC measure less than $1$\% for the worst case over the population. Second, one can replace exact computation of the Wasserstein distance with the computation-efficient approximations, making use of the entropic regularization \cite{Peyre2019} or tree metrics \cite{Le2019}. Third, while it is not affecting computational complexity, the computation of the connectivity matrix entries is independent of each other, so is amenable to the distributed and parallel computation.



The key suggestion that Basti et al. \cite{Basti2020} make is that multivariate connectivity analysis benefits from increased functional inhomogeneity of time series constituting a ROI. The corresponding analysis of the dependence of the inhomogeneity and the performance of the classifiers using Wasserstein distance as connectivity measure relative to the univariate methods based on average and principal component analysis-based representatives is left for the future work.

\vspace{-0.25em}

\subsubsection*{Conclusions}
\vspace{-0.25em}
We have proposed the new Wasserstein distance-based measure of the multivariate brain functional connectivity and demonstrated that it is preferable over the commonly used univariate and multivariate connectivity measures given by the Pearson and distance correlations respectively for the task of autism diagnosis.

%% file: sections/ethics.tex
\section{Compliance with Ethical Standards}

This is a numerical simulation study conducted retrospectively using human subject data made available by Di Martino et al. \cite{DiMartino2014}  for which no ethical approval was required.

\section{Acknowledgements}

No funding was received for conducting this study. The authors have no relevant financial or non-financial interests to disclose.